%% file: main.tex
\documentclass[10pt,twocolumn,letterpaper]{article}

\usepackage[camera-ready]{iccv}      % To produce the REVIEW version

\definecolor{iccvblue}{rgb}{0.21,0.49,0.74}
\usepackage[pagebackref,breaklinks,colorlinks,allcolors=iccvblue]{hyperref}

\def\paperID{12772} % *** Enter the Paper ID here
\def\confName{ICCV}
\def\confYear{2025}

\title{Fish2Mesh Transformer: 3D Human Mesh Recovery from Egocentric Vision}

\author{David C. Jeong, Aditya Puranik, James Vong, Vrushabh Abhijit Deogirikar\\
Ryan Fell, Julianna Dietrich, Maria Kyrarini, Christopher Kitts\\
Santa Clara University\\
Santa Clara, CA\\
{\tt\small [dcjeong, aspuranik, jvong, vdeogirikar, rfell, jdietrich, mkyrarini, ckitts] @scu.edu
}
}
\begin{document}
% \maketitle

\twocolumn[{%
\renewcommand\twocolumn[1][]{#1}%
\maketitle
\begin{center}
    \centering
    \captionsetup{type=figure*}
    \includegraphics[width=1\textwidth]{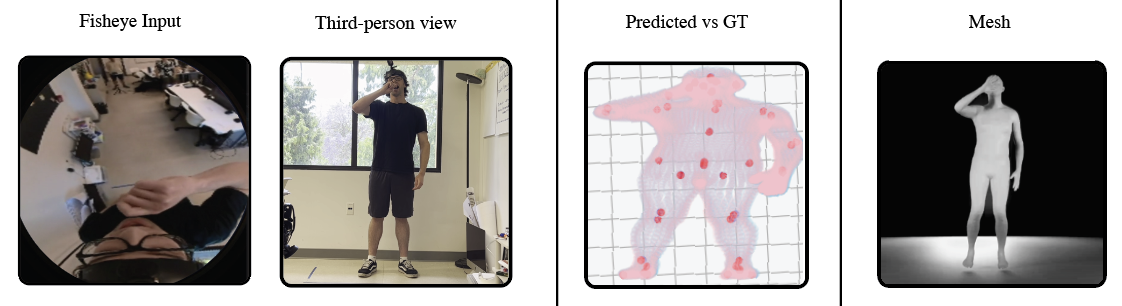}
    \captionof{figure}{The Fish2Mesh pipeline enables accurate 3D mesh recovery from egocentric fisheye perspectives. From left to right: (1) Fisheye Input with a wide field of view; (2) Third-person view (shown for context, not used as input); (3) Predicted (blue) vs. Ground Truth (red) vertices demonstrating near-complete overlap; and (4) Reconstructed 3D human mesh model from the fisheye input. }
    \label{summary}
\end{center}%
}]

% Force the figure to span both columns and appear right after the title using \afterpage
% \afterpage{
%   \begin{figure*}[t]  % 't' for top of the page; 'figure*' spans both columns
%     \centering
%     \includegraphics[width=\textwidth]{fig/summary-figure.png}  % Adjust the size to fit
%     \caption{}
%     \label{summary}
%   \end{figure*}
% }

%%%%%%%%% ABSTRACT
\begin{abstract}
Egocentric human body estimation allows for the inference of user body pose and shape from a wearable camera's first-person perspective. Although research has used pose estimation techniques to overcome self-occlusions and image distortions caused by head-mounted fisheye images, similar advances in 3D human mesh recovery (HMR) techniques have been limited. We introduce \textbf{Fish2Mesh}, a fisheye-aware transformer-based model designed for 3D egocentric human mesh recovery. We propose an egocentric position embedding block to generate an ego-specific position table for the Swin Transformer to reduce fisheye image distortion. Our model utilizes multi-task heads for SMPL parametric regression and camera translations, estimating 3D and 2D joints as auxiliary loss to support model training. To address the scarcity of egocentric camera data, we create a training dataset by employing the pre-trained 4D-Human model and third-person cameras for weak supervision. Our experiments demonstrate that Fish2Mesh outperforms previous state-of-the-art 3D HMR models. Code and data are available on our \href{https://fish2mesh.github.io/}{website}.

\end{abstract}

%%%%%%%%% BODY TEXT
\section{Introduction}

Egocentric human body estimation is an emerging subfield of computer vision~\cite{grauman2024ego,grauman2022ego4d,akada20243d} that focuses on estimating the human body’s pose and shape from a first-person point of view through wearable cameras~\cite{liu2023egofish3d,wang2023scene,xu2019mo,liu2023egohmr,goel2023humans}, typically mounted on the head~\cite{xu2019mo}. Egocentric estimation uses captured images from this unique perspective, enabling applications in assistive robotics~\cite{zhang2013egocentric,kutbi2023egocentric,marina2022head}, augmented reality~\cite{xu2024improving, chalasani2018egocentric, karakostas2024real}, and virtual reality~\cite{nguyen2016recognition}.

Unlike forward-facing cameras~\cite{jiang2021egocentric, Li_2023_CVPR} which excel at egocentric views ideal for third-person reconstruction but are limited for head- and arm-level scenes for first-person view, downward-facing cameras~\cite{xu2019mo, liu2023egohmr} capture a broader, ground-oriented perspective of the human body and its surroundings. This downward-facing perspective enriches contextual detail and boosts the processing potential. Building on these advantages, our work employs head-mounted cameras that capture downward views for human body estimation through key joint mapping, enabling accurate modeling of the body's structure and movements. These benefits motivate the exploration of advanced modeling techniques, such as pose estimation and human mesh recovery, offering promising avenues for capturing precise body structures from egocentric views.

Recent research in human body estimation has focused on pose estimation, which involves predicting the positions of key body joints~\cite{wang2024egocentric}. In contrast, our current work employs \textbf{human mesh recovery} (HMR), or human mesh reconstruction, to enhance spatial precision by mapping egocentric image data onto humanoid 3D mesh models. In addition to meeting the precision demands of realistic simulations and capturing body shape and volume, HMR is better suited to address inconsistencies arising from varying keypoint standards across pose estimation datasets~\cite{liu2023egofish3d,wang2023scene,xu2019mo}. For instance, datasets such as COCO 2017 (32 keypoints)~\cite{lin2014microsoft} and Human3.6M (17 keypoints)~\cite{ionescu2013human3} use different sets of key points to define human poses.

Egocentric HMR faces several unique challenges. First, diverse and annotated egocentric datasets are scarce, as capturing such data requires wearable cameras~\cite{liu2023egohmr}. Second, head-mounted cameras often use fisheye lenses~\cite{liu2023egofish3d,wang2023scene,xu2019mo,liu2023egohmr,goel2023humans}, which introduce distortions, particularly near the edges of the frame. The amount of distortion is dependent on the unique fisheye lens used \cite{grauman2024ego}, which also adds inconsistent standards between datasets and complicates the model training of depth estimation needed for accurate body meshes. Third, the egocentric perspective introduces self-occlusions \cite{zhao2021egoglass}, where farther body parts such as legs can often be covered up. For example, hands and arms frequently block the view of the torso, while the head naturally limits the visibility of the lower body. Additionally, since egocentric cameras are typically worn on the head, they have a restricted field of view, often resulting in partial or complete exclusion of body parts from the frame. Legs, feet, and even parts of the arms may be out of view, especially during dynamic activities such as walking, running, or interacting with objects. Together, these challenges underscore the need for innovative approaches in egocentric HMR to effectively manage limited data, occlusions, head movements, and varying lens distortions.

% Furthermore, head-mounted cameras predominantly employ fisheye lenses ~\cite{liu2023egofish3d,wang2023scene,xu2019mo,liu2023egohmr,goel2023humans} to maximize field of view, an essential requirement given their proximity to the body. However, these lenses introduce significant spherical distortions that fundamentally challenge traditional mesh recovery approaches, particularly in accurate depth estimation. While existing methods attempt to correct these distortions through various projection techniques—often at the cost of spatial accuracy.

In order to address these challenges, we propose a fundamentally different approach in Fish2Mesh, a transformer-based architecture that advances the handling of fisheye distortions in egocentric vision through principled geometric understanding, addressing key limitations in existing methods. Rather than treating these distortions as aberrations that must be corrected, our architecture implements a parameterized \textbf{Egocentric Position Embedding} (EPE), which employs a learnable table that encodes equirectangular geometric information directly. The values in this table represent relative 3D spherical information which is then embedded in the 2D distorted fisheye images.

% Standard transformer position embeddings typically use fixed sinusoidal functions or learnable embeddings that assume a regular grid~\cite{liu2021Swin}, which do not explicitly account for the non-linear distortions introduced by fisheye lenses in egocentric images. Our EPE is designed to capture these subtle, non-linear distortions, thereby facilitating a more accurate mapping between 2D distorted inputs and their corresponding 3D representations. While previous literature on position embeddings does not directly address this specific challenge, our approach is motivated by the need to bridge this gap and is supported by our empirical improvements in egocentric human mesh recovery.

While standard transformer position embeddings typically use fixed sinusoidal functions or learnable embeddings that assume a regular grid~\cite{liu2021Swin} and thus do not account for the non-linear distortions introduced by egocentric fisheye lens images, our EPE is designed to capture these subtle, non-linear distortions, thereby facilitating a more accurate mapping between 2D distorted inputs and their corresponding 3D representations. Although previous literature on position embeddings does not directly address this specific challenge, our approach is motivated by the need to bridge this gap and is supported by our empirical improvements in egocentric human mesh recovery.

In addition to our novel position embedding, we introduce a multi-headed architecture that jointly optimizes SMPL parameters and camera transformations. By implementing auxiliary 3D-to-2D consistency objectives alongside parameter regression, we create a geometric framework that maintains spatial coherence across both 2D and 3D dimensions, which is critical for egocentric perspectives.

%we enhance the egocentric data landscape through
To address the limitations of existing datasets, we augment an existing dataset by introducing a prompt-based collection system that captures natural human motion. Moving beyond scripted actions, our approach generates a diverse range of movements that include realistic head motions and self-occlusions that are characteristic of daily activities. This improved data foundation, combined with our geometric-aware architecture, enables robust performance in challenging real-world scenarios.

In this paper, we propose the following novel contributions in the pursuit of egocentric HMR:
\begin{itemize}
    \item \textbf{Fish2Mesh:} A fisheye-aware, transformer-based architecture that takes egocentric fisheye image inputs and reconstructs 3D human meshes over more conventional joint estimations.
     \item \textbf{Egocentric Position Embedding:} A novel equirectangular projection method with learned position embedding to estimate the 3D spherical coordinates of each fisheye image pixel to address image distortion. 
    % \item \textbf{Multi-task heads for SMPL} parametric regression and camera transformations to regress the 3D and 2D coordinates as auxiliary losses for egocentric human mesh reconstruction.
    % \item We provide both quantitative and qualitative results on various datasets to demonstrate the effectiveness of our proposed model.
    \item \textbf{Dataset augmentation:} A prompt-based data collection method that captures naturalistic and real-world human movements with real-world occlusions, significantly improving model performance in practical applications. 
    % \item  \textbf{Comprehensive experimental evaluations} across multiple egocentric datasets, demonstrating the effectiveness of our proposed model through both quantitative and qualitative analysis. 
\end{itemize}

 % Furthermore, most head-mounted cameras use a fisheye lens to capture a large field of view due to their proximity to the body, but such perspectives introduce significant distortion in the images, hampering depth estimation, a critical component of accurate human mesh recovery.

% Our Contributions
% we propose a Fish-eye aware transformer-based model, called FisheyeFormer to reduce the distortion of fisheye images as the input data.

% We propose multi-task heads for SMPL parametric-regression and camera transformations to regress the 3D and 2D cooridnate as the auxilliant loss for egobody human mesh reconstruction.

% Quantitative and qualitative results on different datasets demonstrate the effectiveness of our proposed model.

\section{Related Work}

\begin{figure*}[t!]
\centering
\includegraphics[width=.90\textwidth]{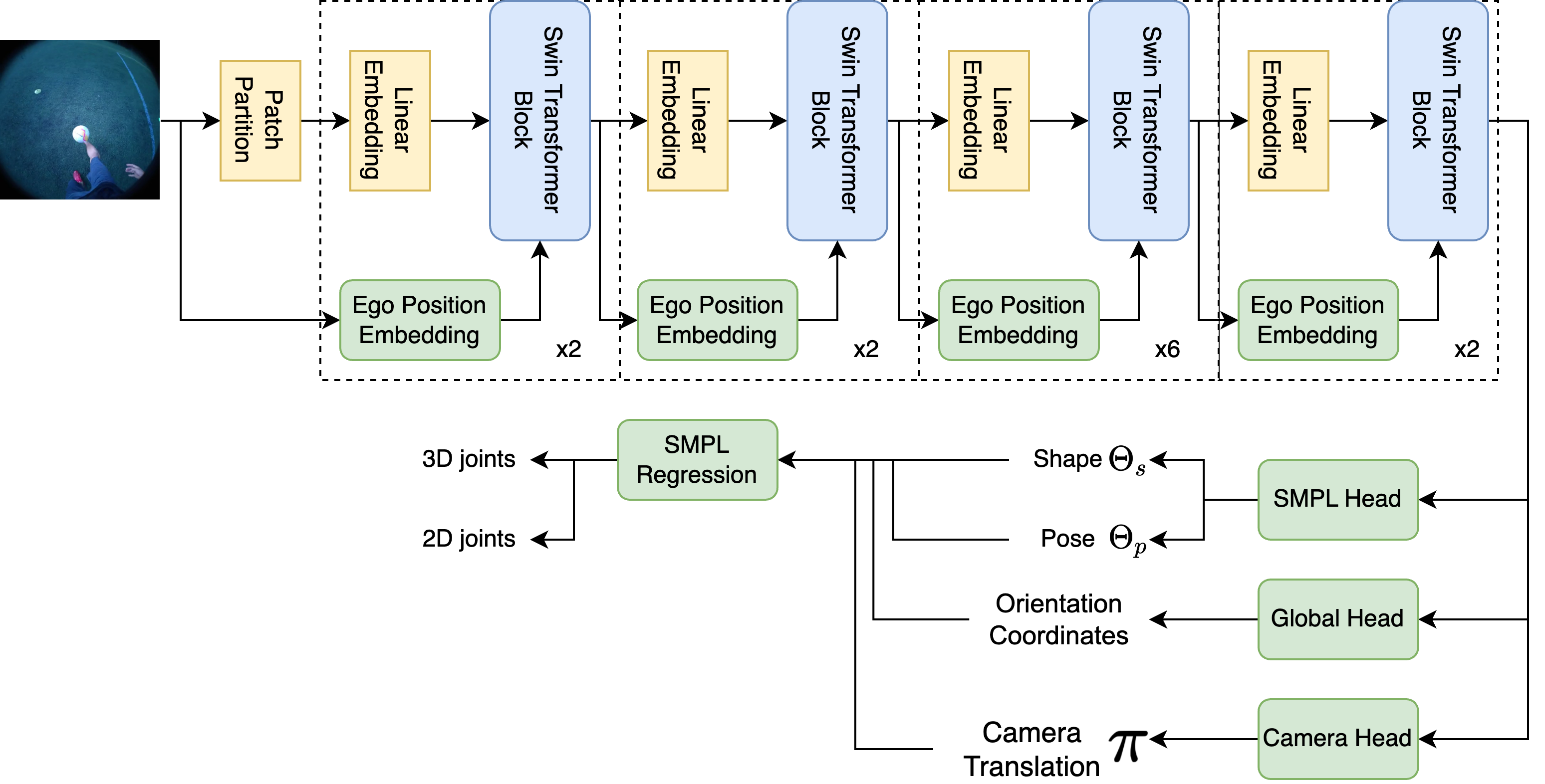} % Reduce the figure size so that it is slightly narrower than the column.
\caption{The architecture of the Fish2Mesh transformer model. The SMPL parameters $\Theta_{s}, \Theta_{p}$ are calculated to recover the human mesh, where $\Theta_{s}$ and $ \Theta_{p}$ refer to the shape parameters and pose parameters respectively. 
%$GT_{\Theta_{s}}, GT_{\Theta_{p}}, GT_{orient}, GT_{3D}$ and $ GT_{2D}$ are the ground truth from the datasets.
}
\label{frame}
\end{figure*}

\subsection{Egocentric Human Body Estimation}

\subsubsection{Egocentric Pose Estimation}
% more research interests this fields and more challenge than 3rd person veiw
% Early work ~\cite{rhodin2016egocap,zhao2021egoglass,akada2022unrealego} using stereo cameras which is more challenge
Research in egocentric human body estimation has gained considerable momentum, building on early pose estimation works that used stereo (i.e., dual) camera systems, such as EgoCap~\cite{rhodin2016egocap}, EgoGlass~\cite{zhao2021egoglass}, and UnrealEgo~\cite{akada2022unrealego}. However, such stereo-egocentric vision systems were constrained by the alignment and calibration of stereo cameras, the computational overhead required to process stereo imagery, and the physical burden of additional hardware. As such, egocentric body estimation shifted towards monocular (i.e., single-camera) systems~\cite{liu2023egofish3d,liu2023egohmr,tome2020selfpose,xu2019mo}, which simplify hardware complexity and reduce user burden while addressing the inherent challenges of egocentric vision systems. The estimation of monocular egocentric pose was further improved by the development of new training data sets~\cite{tome2020selfpose,xu2019mo}, such as SelfPose~\cite{tome2020selfpose} and Mo2Cap2~\cite{xu2019mo}, thus eliminating the need for stereo camera setup, reducing complexity and improving usability.

\subsubsection{Egocentric Human Mesh Recovery}
While egocentric human body estimation research has focused mainly on estimating joints through pose estimation~\cite{liu2023egofish3d,wang2023scene,xu2019mo}, EgoBody~\cite{zhang2022egobody} and EgoMoCap~\cite{liu20214d} first sought to introduce the challenge of addressing human mesh recovery using original egocentric image datasets. However, EgoBody and EgoMoCap reconstructed the human mesh of a confederate interlocutor that was the subject of the egocentric image data from the first-person view of the wearer. In other words, EgoBody and EgoMoCap achieved third-person human mesh recovery of egocentric videos, which was a departure from the first-person body estimation achieved via egocentric pose estimators like EgoCap~\cite{rhodin2016egocap} and EgoGlass~\cite{zhao2021egoglass}.

% EgoHMR~\cite{liu2023egohmr} is the only paper solve the dataset and propose a HMR model.
% But it has servel disadvantages:
% --- diffusion model is a generate model, whose output is different which is not suitable for many applications
% --- diffusion model is a very larger model,which cannot apply on real-time system.
% --- this model suffer from the distrotion of fisheye images
\subsubsection{State of the Art: EgoHMR}
In perhaps the most significant contribution to egocentric vision to date, EgoHMR~\cite{liu2023egohmr} proposed a diffusion model to recover a 3D human mesh from the camera user using an egocentric video dataset with SMPL labels~\cite{loper2023smpl}. Despite its strengths, the inherently variable nature of EgoHMR's diffusion model makes it unsuitable for applications requiring high precision, such as real-time XR systems \cite{xu2024improving, chalasani2018egocentric, karakostas2024real} and assistive human-robot interaction \cite{zhang2013egocentric,kutbi2023egocentric,marina2022head}. More critically, the EgoHMR model fails to address fisheye image distortions, which can degrade mesh recovery quality. In spite of EgoHMR's significant advances, the above limitations underscore the need for further human mesh recovery methods that can meet the rigorous demands of egocentric vision with greater resilience to fisheye images.

\subsection{Egocentric Vision Transformer Models}
% Introduce deep learning vision model on egocentric images
% --- what's the challenge and different compared with 3rd person's view
%(fisheye lens distortions)
One of the significant challenges in developing deep learning models for egocentric vision is that the camera's close proximity to the body introduces image distortions that complicate tasks like feature extractions and pose estimation ~\cite{liu2023egofish3d,liu2023egohmr,tome2020selfpose,xu2019mo}. Panoformer~\cite{shen2022panoformer} attempts to address fisheye distortion challenges by estimating depth from panoramic images using tangent patches from the spherical domain, effectively correcting distortions. However, the local attention mechanisms used in Panoformer have a restricted receptive field, which limits their ability to capture long-range dependencies and global context effectively. Alternatively, EGformer~\cite{yun2023egformer} introduces equirectangular-aware, multi-head self-attention to enlarge the receptive field along the horizontal and vertical axes, but this comes at a higher computational cost and a greater number of network parameters, reducing efficiency. More recently, FisheyeViT~\cite{wang2024egocentric} employs a patch-based strategy, extracting many undistorted patches from fisheye images and fitting them into a transformer network. However, the FisheyeViT method relies on small patch sizes, leading to a larger number of patches and thus significant preprocessing time. 

Our proposed model introduces a novel \textbf{position embedding tailored for the Swin Transformer} that allows the model to capture both horizontal and vertical information more effectively in the equirectangular domain.

% In egocentric images, the field of view is constrained to what the wearer can see, leading to issues such as severe distortions due to the proximity of the camera to the subject, such as fisheye lens distortions, which can further complicate feature extraction and pose estimation~\cite{liu2023egofish3d,liu2023egohmr,tome2020selfpose,xu2019mo}.

% There are serval models~\cite{wang2024egocentric,shen2022panoformer,yun2023egformer} to reduce fisheye lens distortions.
% Itro Panoformer~\cite{shen2022panoformer} pros and cons. 
% Egoformer~\cite{yun2023egformer} pros and cons.
% FisheyeVit~\cite{wang2024egocentric} pros and cons. 
% --- we wanna propose a way didnot cost lots of time, but can efficiently reduce distrotion
% Thus, in this paper we propose a new egocentic position embedding which let Swin Transformer model aware the equirectangular horizontal and vertical information.

% FisheyeViT~\cite{wang2024egocentric} takes a different approach by gridding the input data, extracting undistorted patches from the fisheye image, and fitting these patches as tokens into a transformer network. To address these issues, our paper proposes a new position embedding for the Swin Transformer model, enabling it to better capture equirectangular horizontal and vertical information while maintaining computational efficiency.

\section{Methodology}
\subsection{Problem Statement}
The primary objective of this paper is to achieve accurate HMR from egocentric RGB images captured by a head-mounted fisheye camera. We denote the captured image at each frame as $I$, where each pixel is defined by its 2D coordinates $x_{2D}$ and $y_{2D}$, forming the input data of our model. As illustrated in Fig.~\ref{frame}, our proposed \textit{Fish2Mesh} framework consists of three components: \textbf{(1) Egocentric Position Embedding} to mitigate fisheye distortion, \textbf{(2) Swin Transformer blocks}~\cite{liu2021Swin} for hierarchical feature extraction, and \textbf{(3) Task-specific heads} for predicting SMPL parameters as output. The outputs of Fish2Mesh include the SMPL parameters $\Theta_{s}$ for body shape, $\Theta_{p}$ for body pose, camera transformation $\Pi$, global orientation $L$, 3D key joints $L_{3D}$ and 2D key joints $L_{2D}$. The positions of 3D body joints can be directly derived from the SMPL regression model. 

%which is integral for egocentric human pose estimation.

\subsection{Egocentric Position Embedding}
\begin{figure}[t!]
\centering
\includegraphics[width=.50\textwidth]{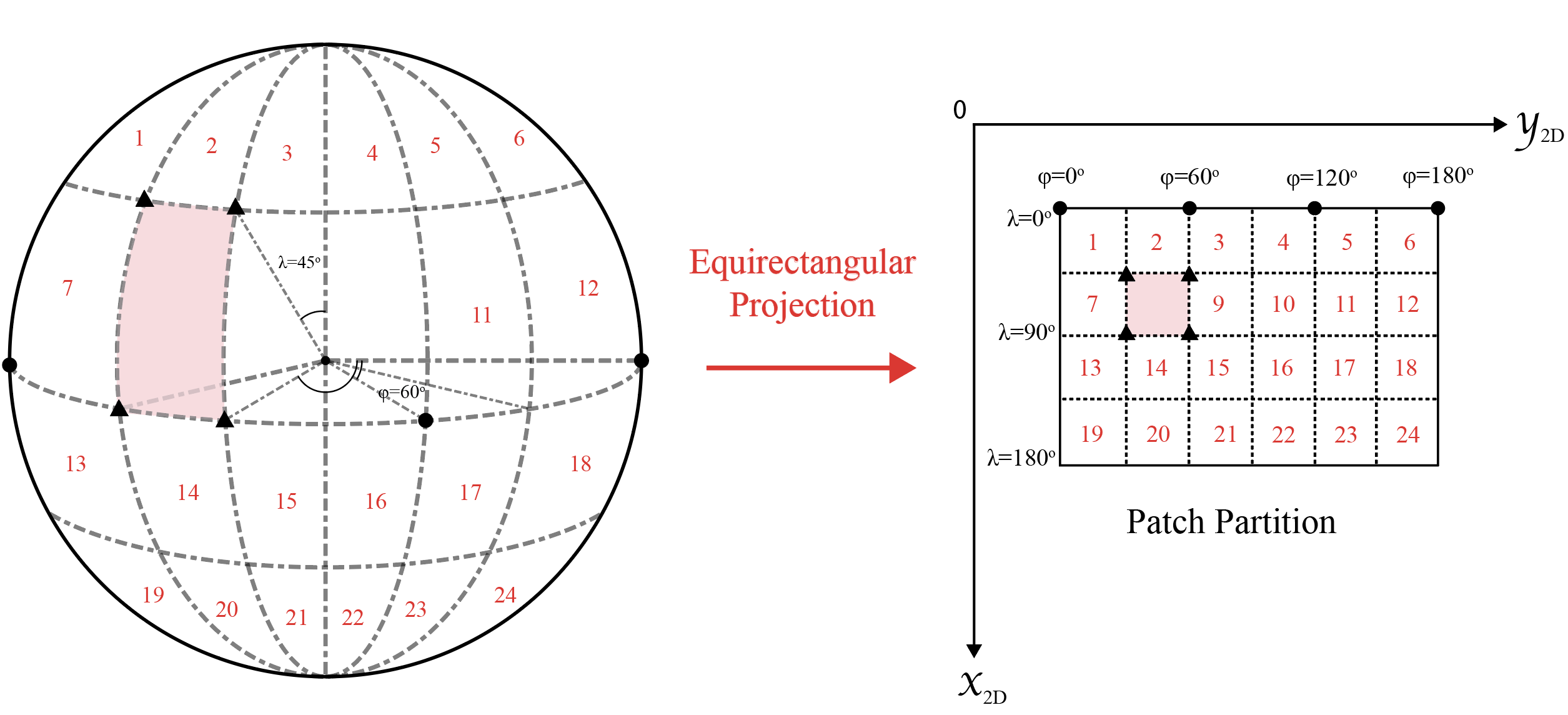} % Reduce the figure size so that it is slightly narrower than the column.
\caption{Equirectangular projection from spherical image.}
\label{position}
\end{figure}

% egocentric images/ fisheye images are different from regular camera's images
% --- raw data of fisheye images are on the sphere surface
% there are distortion by projection in preious works.
% in this work, we propose egocentric postion embedding to reduce the distrotion and without complexity computation
Unlike conventional images, fisheye images are projected onto a spherical surface, which introduces image distortions as shown in Fig.~\ref{position}. Previous approaches~\cite{liu2023egofish3d,wang2023scene,xu2019mo,liu2023egohmr,goel2023humans} often unintentionally amplify these distortions through projection techniques, leading to a further loss in spatial accuracy. To mitigate this, we propose a parameterized Egocentric Position Embedding that applies \textbf{equirectangular geometry bias} directly to the raw image elements. This embedding method reduces distortion by adapting to the fisheye images' inherent spherical nature while avoiding computational complexity from traditional correction methods. This equirectangular geometry bias also adapts to the unique spatial constraints of egocentric data, where the camera’s proximity to the wearer introduces further challenges that standard positional embeddings struggle to address. Our Vision Transformer-inspired  approach~\cite{dosovitskiy2020image} also introduces positional information to transformer blocks to effectively retain 3D spatial context. 

\subsubsection{Equirectangular Projection}
 % A 180-degree fisheye image is constructed by projecting the spherical image onto a 2D plane using the following formulas:
As seen in Fig \ref{position}, we project the spherical 180-degree fisheye image onto a 2D plane to construct an equirectangular panoramic image using the following formula:
    \begin{align}
    x_{2D} &= R\left(\lambda - \lambda_0\right) \cos \varphi_1 \\
    y_{2D} &= R\left(\varphi - \varphi_0\right) \\
    \lambda &= \frac{x_{2D}}{R \cos \varphi_1} + \lambda_0 \\
    \varphi &= \frac{y_{2D}}{R} + \varphi_0
    \end{align}
where $R$ is the radius of the spherical projection and depends on the physical lens properties of the camera, and $x_{2D}$ and $y_{2D}$ are the coordinates of input images. 

$\varphi_1$ represents a reference latitude on the spherical projection. It is used to adjust the scaling factor in the transformation, ensuring that the projection accurately maps the spherical coordinates onto the 2D plane. This parameter helps maintain the correct aspect ratio and spatial alignment when converting from spherical to equirectangular representation. $\lambda_0$ is the reference longitude of the spherical coordinate system that sets the origin of the longitude measurement, serving as the reference point for horizontal positioning in the equirectangular projection. This ensures that the panoramic image is correctly centered when unwrapped onto the 2D plane. $\varphi_0$ is the reference latitude, which determines the vertical center of the projection. It specifies the latitude where the unwrapping process begins, helping to align the image correctly in the 2D coordinate system. 

The variables $\lambda \in(0, \pi)$ and $\varphi\in(0, \pi)$ denote the longitude and latitude, respectively. However, since our position embedding is implemented as a learnable table that requires discrete indices, these continuous values cannot be used directly. We restrict $\varphi$ to the range $(0, \pi)$ because our fisheye image captures a 180 degree field of view.

% FFmpeg is a collection of libraries and tools to process multimedia content such as audio, video, subtitles and related metadata.

%rather than a full 360 degrees.

To recover the 3D spatial context, we convert the equirectangular images back to spherical coordinates:
    \begin{equation}
    \left\{\begin{array}{l}
    x_{3 D}=R \cdot \sin (\varphi) \cdot \cos (\lambda) \\
    y_{3 D}=R \cdot \sin (\varphi) \cdot \sin (\lambda) \\
    z_{3 D}=R \cdot \cos (\varphi) 
    \end{array}\right.
    \end{equation}
This back-projection is essential for reconstructing the 3D spatial properties of the original scene, ensuring the model retains an accurate understanding of depth and orientation.
    
The final transformation extends this logic: 
\begin{equation}
\left\{\begin{array}{l}
x_{3 D}=R \cdot \sin \left(\frac{y_{2 D}}{R}+\varphi_0\right) \cdot \cos \left(\frac{x_{2 D}}{R \cos \varphi_1}+\lambda_0\right) \\
y_{3 D}=R \cdot \sin \left(\frac{y_{2 D}}{R}+\varphi_0\right) \cdot \sin\left(\frac{ x_{2 D}}{R \cos \varphi_1}+\lambda_0\right) \\
z_{3 D}=R \cos \left(\frac{y_{2 D}}{R}+\varphi_0\right)
\end{array}\right.
\end{equation}

The resulting $x_{3D}$, $y_{3D}$, and $z_{3D}$ coordinates provide a rich, structured representation of the spherical distortion. By discretizing these transformed coordinates, we create a set of indices for our learnable 3D position embeddings $POS[x_{3D}, y_{3D} ,z_{3D}]$. Unlike standard position embeddings that use fixed sinusoidal functions or direct spherical coordinates, our approach embeds discretized 3D coordinates into the Swin Transformer blocks, enabling the model to capture the spherical characteristics of the fisheye input and learn complex distortions while preserving essential 3D spatial information.

\subsection{Swin Transformer Blocks}
Our Fish2Mesh model employs the Swin Transformer~\cite{liu2021Swin} architecture as its backbone, utilizing its \textbf{patch merging layers} as the network deepens, prior to the four main Swin blocks shown in Fig~\ref{frame}. We introduce our Egocentric Position Embedding $POS[x_{3D}, y_{3D} ,z_{3D}]$ of spherical information to the equirectangular images, and feed them into the patch merging layers. In the initial patch merging mechanism, we hierarchically consolidate the feature representations by merging adjacent patches. Specifically, we fuse features from a $2\times2$ patch grid into a $4C$ vector, where $C$ is the channel depth, then linearly map it via $W: \mathbb{R}^{4C} \to \mathbb{R}^{2C} \times (H/2, W/2)$, which halves the space and doubles the channels~\cite{liu2021Swin}. An embedding layer casts this into Swin tokens, priming EPE for fisheye’s wild curves.

% Specifically, we aggregate features from 2 × 2 contiguous patch regions, concatenating their representations to form a unified feature vector of dimensionality 4C, where C denotes the initial channel dimension. This concatenated representation undergoes a linear transformation that simultaneously reduces spatial resolution by a factor of 4, while doubling the channel dimension to 2C. Subsequently, the transformed features are processed through a linear embedding layer, which projects them into the token space required for the downstream Swin Transformer blocks. 

% In the initial patch merging layer, the features from each group of 2 × 2 neighboring patches are concatenated, resulting in a feature vector with a dimension of 4C, where C is the number of channels. A linear layer is then applied, reducing the token count by a factor of 4, and the output dimension is set to 2C.  After this transformation, the intermediate feature is passed through a linear embedding layer which converts the data into the 2D tokens required as input for the Swin Transformer blocks.

The two successive Swin Transformer blocks consist of a window-based and a shifted window-based multi-head self-attention (MSA) module. The window-based MSA module operates within fixed, non-overlapping windows, while the shifted window-based MSA module in the subsequent block introduces connections between neighboring windows from the previous layer, enabling the model to capture both local and long-range dependencies. Each MSA module is followed by a two-layer MLP with GELU nonlinearity, and a LayerNorm (LN) is applied before each MSA and MLP. Residual connections are added after each module to enhance gradient flow and stabilize training.

We remove the relative position bias in the transformer blocks because our features already contain positional information, making the addition of relative position bias redundant. The final attention equation is shown in Formula 7, where \( Q, K, V \in \mathbb{R}^{M^2 \times d} \) represent the query, key, and value matrices, respectively, \( d \) is the dimension of the query and key, and \( M^2 \) is the number of patches within a window.
\begin{equation}
\operatorname{Attention}(Q, K, V)=\operatorname{softmax}\left(Q K^T / \sqrt{d}\right) V
\end{equation}

\subsection{Task-Specific Heads and Losses}
After extracting key features from the Swin Transformer blocks, we employ three task heads for SMPL parameters prediction, camera parameters, and global orientation coordinates. The training of our Fish2Mesh model relies on a well structured loss function that integrates SMPL shape $\Theta_{s}$ losses, SMPL pose $\Theta_{p}$ losses, and orientation losses, along with auxiliary 3D and 2D losses shown in Formula 8.

The total loss function is defined as:
\begin{align}
\mathcal{L} &= a * (\mathcal{L}_{\text{SMPL}} + \mathcal{L}_{\text{orient}}) + b * \mathcal{L}_{\text{3D}} + c * \mathcal{L}_{\text{2D}}\\
\mathcal{L}_{\text{SMPL}} &= \left\|GT_{\Theta_{s}} - \Theta_{s}\right\|_2^2 + \left\|GT_{\Theta_{p}} - \Theta_{p}\right\|_2^2 \\
\mathcal{L}_{\text{orient}} &= \left\|GT_O - O\right\|_1 \\
\mathcal{L}_{\text{3D}} &= \left\|GT_{3D} - X\right\|_1 \\
\mathcal{L}_{\mathrm{2D}} &= \left\|GT_{2D} - \pi(X)\right\|_1
\end{align}

where \(\mathcal{L}\) is the total loss function used for model training and \(a\), \(b\), and \(c\) are weights assigned to the different loss components. \(\mathcal{L}_{\text{SMPL}}\) represents the SMPL parameter loss, which includes shape loss \(\Theta_{s}\) and pose loss \(\Theta_{p}\), and \(GT_{\Theta_{s}}\) and \(GT_{\Theta_{p}}\) are ground truth SMPL shape and pose parameters, respectively. The orientation loss \(\mathcal{L}_{\text{orient}}\) ensures that the predicted orientation \(O\) matches the ground truth  \(GT_O\) using the L1-based loss, \(\mathcal{L}_{\text{3D}}\) is the auxiliary 3D loss with \(GT_{3D}\) as the ground truth 3D coordinates, and \(\mathcal{L}_{\mathrm{2D}}\) is the auxiliary 2D loss with \(GT_{2D}\) as the ground truth 2D coordinates. \(X\) denotes the predicted 3D coordinates, and \(\pi(X)\) is \(X\)'s projection onto the 2D plane.

This multi-loss approach ensures that the model optimizes 3D reconstructions and the alignment of the 2D projections with ground truth. Balancing between 3D and 2D consistency allows the model to converge more reliably, enhancing both accuracy and generalization. By carefully weighing the different loss components, our robust training process strikes a balance between optimizing the SMPL parameters, refining the global orientation, and improving overall mesh recovery performance. We train our Fish2Mesh model from scratch using loss function (8) in an end-to-end manner.

%(harrison could you help me to summarize all information in doc for this paper)

\begin{table*}[!h]
\centering
\tiny % Reduce the font size
\caption{Evaluation results across 3 datasets, including MPJPE, MPVPE, PA-MPJPE, and PA-MPVPE (all in mm).}
\label{table1}
\vspace{-0.4cm}
%columnwidth
\resizebox{14cm}{!}{%
\begin{tabular}{@{}cccccccc@{}}
\toprule
Model                              & MPJPE & MPVPE & PA-MPJPE & PA-MPVPE & Dataset \\ \midrule
4DHuman~\cite{liu20214d}            & 390.037 & 521.349 & 90.037  & 129.849  & ECHP    \\
FisheyeViT~\cite{wang2024egocentric}   & 4594.004 & /      & 94.184  & /        & ECHP    \\
EgoHMR~\cite{liu2023egohmr}         & 84.332  & 99.983  & 64.112  & 79.031   & ECHP    \\
\multicolumn{1}{l}{Fish2Mesh(ours)} & \textbf{79.699} & \textbf{98.111} & \textbf{57.671} & \textbf{75.322} & ECHP    \\ \midrule
4DHuman~\cite{liu20214d}            & 320.005 & 402.222 & 120.305 & 132.832  & Ego4D   \\
FisheyeViT~\cite{wang2024egocentric}   & 491.975 & /      & 91.975  & /        & Ego4D   \\
EgoHMR~\cite{liu2023egohmr}         & 224.423 & 311.129 & 114.423 & 128.999  & Ego4D   \\
Fish2Mesh(ours)                     & \textbf{71.934}  & \textbf{84.116}  & \textbf{41.931}  & \textbf{54.756}  & Ego4D   \\ \midrule
4DHuman~\cite{liu20214d}            & 298.233 & 320.944 & 98.613  & 120.304  & Our     \\
FisheyeViT~\cite{wang2024egocentric}   & 493.117 & /      & 93.547  & /        & Our     \\
EgoHMR~\cite{liu2023egohmr}         & 227.552 & 294.484 & 127.55  & 144.184  & Our     \\
Fish2Mesh(ours)                     & \textbf{57.352}  & \textbf{71.233}  & \textbf{37.242}  & \textbf{51.58}   & Our     \\ \bottomrule
\end{tabular}%
}
\end{table*}

\section{Experimental Results}

\subsection{Datasets}
To comprehensively evaluate the performance of our Fish2Mesh model, we utilized three distinct datasets: Ego4D~\cite{grauman2022ego4d}, ECHP~\cite{liu2023egofish3d}, and an \textbf{enhanced ECHP dataset} using prompt-based collection that captures natural movements and real-world occlusions. For further details about experimental setup and enhancements using prompts, please refer to the supplementary material. Each of these datasets offers unique characteristics and contributions that enhance the diversity and robustness of our model.

% introduce Ego4D
\subsubsection{Ego4D Dataset}

The Ego4D dataset is an expansive dataset that captures a wide range of egocentric interactions. Comprising of 3,670 hours of video footage from 923 participants across nine countries and 74 locations, this dataset serves as a cornerstone for research in first-person visual perception. Its richness in capturing various daily activities and diverse environments ensures that our model is trained and tested on a broad spectrum of real-world interactions, enhancing its generalization to unseen scenarios.

% introduce ECHP
\subsubsection{ECHP Dataset}

The ECHP dataset is tailored specifically for egocentric 3D human pose estimation. It uses a GoPro camera with a fisheye lens to capture daily human actions in varied environments. The ECHP dataset features a multi-camera setup and includes ground truth annotations, allowing for precise evaluation of pose estimation models. This dataset is invaluable for assessing the accuracy of our model in real-world scenarios, where precise 3D reconstructions are critical.

\subsubsection{Expanding The ECHP Dataset}
To address the scarcity of high-quality egocentric camera data, we adopt the EgoCentric-Human-Pose (ECHP) Dataset~\cite{liu2023egofish3d} for data expansion, leveraging consistent equipment and processing techniques. To improve label accuracy, we utilize the 4DHuman~\cite{liu20214d} model as a self-supervisor rather than PARE~\cite{kocabas2021pare}. Incorporating the 4DHuman model significantly enhances data diversity, as the actions in the existing ECHP dataset are overly simplistic or artificial (e.g., walking actions are simulated by participants merely walking in place). As a result of participants' confinement to a limited space, they cannot exhibit natural head turns or shakes that are naturalistic to everyday human activity. Ultimately, the ECHP data lacks the richness and realism seen in real-life scenarios, where head movements are more varied, leading to common issues like self-occlusion, as illustrated in Fig. \ref{visual}.

To counter these limitations, we introduce a \textbf{prompt-based data collection system}. Instead of directing participants to perform rigid actions, we guide them using open-ended prompts such as "Stirring a Big Pot." By engaging participants in more interactive and imaginative scenarios rather than constrained actions like walking, we aim to produce more natural and dynamic movements. The benefit of this approach is two fold: it helps enhance the realism of the recorded data and increases the variety of head movements, thus leading to improved model training and more accurate human pose estimations. The details of the ECHP extension are shown in the supplement material. 

Additionally, to address challenges like self-occlusions, we use prompts to create situations where these issues are more likely to occur. For example, participants may be asked to perform high-frequency movements that induce self-occlusion or result in parts of the body being outside the frame. This enables us to collect data specifically designed to train and evaluate the model on how well it can overcome these problems.

% \subsubsection{Newly Proposed Dataset}

% In addition to the Ego4D and ECHP datasets, we introduced a new dataset to further enrich the diversity of our training data. This dataset includes videos captured using different types of fisheye cameras, providing varied input perspectives that are crucial for improving the model’s robustness and generalization.

% By integrating these datasets, our approach ensures a comprehensive and diverse training set. The use of four different types of fisheye cameras across these datasets provides varied input perspectives, enhancing the model’s ability to handle different camera settings and environments. The datasets were divided into validation and testing sets, enabling us to evaluate our model alongside state-of-the-art (SOTA) models across separate testing datasets derived from Ego4D, ECHP, and our proposed data.

% training dataset
% validation dataset
% testing dataset

\subsection{Evaluation Metrics}

% PA-MPJPE and PA-MPVPE
For evaluating the performance of our Fish2Mesh Transformer, we employed PA-MPJPE (Procrustes-Aligned Mean Per Joint Position Error) and PA-MPVPE (Procrustes-Aligned Mean Per Vertex Position Error) as our primary evaluation metrics. 
% These metrics were chosen over the traditional MPJPE (Mean Per Joint Position Error) and MPVPE (Mean Per Vertex Position Error) to address the challenge of the different camera coordinates. 
The main reason for this choice is the evaluation of models across different datasets. Previous state-of-the-art (SOTA) models often trained on one dataset and were then evaluated on another, which may not provide a fair comparison. In real-world applications, however, it is not feasible to retrain models on different datasets. Therefore, PA-MPJPE and PA-MPVPE offer a more reliable metric for evaluating the performance of models in varying camera coordinates, as they align the results across datasets and allow for a more accurate comparison of model performance. Nevertheless, we still provide MPJPE and MPVPE for reference to give a comprehensive understanding of the model’s performance on the traditional metrics.

% why we choose MPJPE and MPVPE
\subsubsection{Challenges with Traditional Metrics}

% Traditional metrics like MPJPE and MPVPE face several challenges: 

% \begin{itemize}
%     \item \textbf{Camera Coordinate Frame}: Most methods estimate the 3D pose in the camera coordinate frame rather than the world coordinate frame, leading to variations caused by camera positioning.
%     \item \textbf{Body Orientation}: The body may be tilted relative to the world, complicating the direct assessment of pose accuracy.
%     \item \textbf{Flipping and Symmetry}: Cases where the body is flipped left to right can result in higher errors even if the pose is approximately correct.
% \end{itemize}
% These factors collectively can lead to significant inaccuracies, making MPJPE and MPVPE less reliable for precise pose estimation.

% Traditional metrics such as MPJPE and MPVPE face several inherent challenges \cite{ionescu2013human3}. First, many methods estimate the 3D pose within the camera coordinate frame instead of the world coordinate frame, introducing variations due to the camera's positioning. Additionally, the body’s orientation relative to the world can be misaligned, which complicates the assessment of pose accuracy. Furthermore, cases of flipping and symmetry, such as when the body is flipped left to right, may result in increased error despite the pose being approximately correct. Collectively, these factors can lead to significant inaccuracies, reducing the reliability of MPJPE and MPVPE for precise pose estimation.

Traditional metrics like MPJPE and MPVPE struggle with inherent flaws~\cite{ionescu2013human3}: 3D poses estimated in camera coordinates vary with positioning, body orientation misaligns with the world frame, and inversion (e.g., left-right symmetry) inflates errors despite near-correct poses. These issues, exacerbated by fisheye lens distortions, compromise reliability and motivate our adoption of PA-aligned metrics to fairly assess EPE’s pose accuracy.

To address these challenges, we use PA-MPJPE and PA-MPVPE with Procrustes alignment (PA), which adjusts predicted poses for scale, rotation, and translation ~\cite{liu2023egohmr}. This minimizes external distortions like body tilt—focusing metrics on pose accuracy, enhancing evaluation of our EPE’s 2D-to-3D mapping.

% The PA-MPJPE and PA-MPVPE metrics could be mathematically defined as:

% \[
% \text{PA-MPJPE} = \frac{1}{N} \sum_{i=1}^{N} \left\| P(\hat{J}_i) - J_i \right\|_2
% \]

% \[
% \text{PA-MPVPE} = \frac{1}{N} \sum_{i=1}^{N} \left\| P(\hat{V}_i) - V_i \right\|_2
% \]

% where $N$ denotes the total number of samples in the evaluation dataset, \( P \) denotes the Procrustes alignment function; \( \hat{J}_i \) is the predicted joint position, and \( J_i \) is the ground truth joint position for the \( i \)-th joint; $\hat{V}_i$ is the predicted vertex position, and $V_i$ is the ground truth vertex position for the $i$-th vertex.

% The PA-MPVPE metric could be defined as:

% \[
% \text{PA-MPVPE} = \frac{1}{N} \sum_{i=1}^{N} \left| P(\hat{V}_i) - V_i \right|_2
% \]

% where $N$ denotes the total number of samples in the evaluation dataset, $P$ represents the Procrustes alignment function, $\hat{V}_i$ is the predicted vertex position, and $V_i$ is the ground truth vertex position for the $i$-th vertex.

% By using PA-MPJPE and PA-MPVPE, we achieve a more robust evaluation that focuses on the quality of the pose estimation, independent of irrelevant factors like body orientation and camera perspective.

\subsection{Quantitative Results}
% \usepackage{booktabs}
% \usepackage{graphicx}
% \begin{table}[]
% \caption{Evaluation results of all models across 3 datasets. Both PA-MPJPE and PA-MPVPE are measured in millimeters (mm).}
% \resizebox{\columnwidth}{!}{%
% \begin{tabular}{@{}cccc@{}}
% \toprule
% Model                              & PA-MPJPE & PA-MPVPE & Dataset \\ \midrule
% 4Dhuman~\cite{liu20214d}                            & 90.037   & 129.849  & ECHP    \\
% FisheyeViT~\cite{wang2024egocentric}                            & 94.184   & /        & ECHP    \\
% EgoHMR~\cite{liu2023egohmr}                             & 64.112   & 79.031   & ECHP    \\
% \multicolumn{1}{l}{Fish2Mesh(our)} & \textbf{57.671}   & \textbf{75.322}   & ECHP    \\ \midrule
% 4Dhuman~\cite{liu20214d}                            & 120.305  & 132.832  & Ego4D   \\
% FisheyeViT~\cite{wang2024egocentric}                            & 91.975   & /        & Ego4D   \\
% EgoHMR~\cite{liu2023egohmr}                             & 114.423  & 128.999  & Ego4D   \\
% Fish2Mesh(our)                     & \textbf{41.931}   & \textbf{54.756}   & Ego4D   \\ \midrule
% 4Dhuman~\cite{liu20214d}                            & 98.613   & 120.304  & Our     \\
% FisheyeViT~\cite{wang2024egocentric}                            & 93.547   & /        & Our     \\
% EgoHMR~\cite{liu2023egohmr}                             & 127.55   & 144.184  & Our     \\
% \multicolumn{1}{l}{Fish2Mesh(our)} & \textbf{37.242}   & \textbf{51.58}    & Our     \\ \bottomrule
% \end{tabular}%
% }
% \label{table1}
% \end{table}

The results for PA-MPJPE and PA-MPVPE across the three datasets are summarized in Table~\ref{table1}. Our model outperforms all state-of-the-art (SOTA) models, achieving the best results. Since FisheyeViT~\cite{wang2024egocentric} is a pose estimation method, PA-MPVPE cannot be calculated for it, so we focus on comparing PA-MPJPE values with other SOTA models. Notably, while EgoHMR~\cite{liu2023egohmr} shows overfitting on its own dataset, ECHP, its performance is still 11.2\% higher on PA-MPJPE and 4.9\% higher on PA-MPVPE (lower values indicate better performance). Furthermore, Fish2Mesh outperforms other SOTA models on our extended dataset, demonstrating its superior ability to handle challenges such as self-occlusion and parts of the body being outside the frame. This is because our dataset includes high-frequency data where these issues frequently occur. The testing data in our dataset specifically addresses these two challenges, and we present visual results in the next section to further illustrate the model's performance.

\begin{figure*}[t!]
\centering
\includegraphics[width=0.9\textwidth]{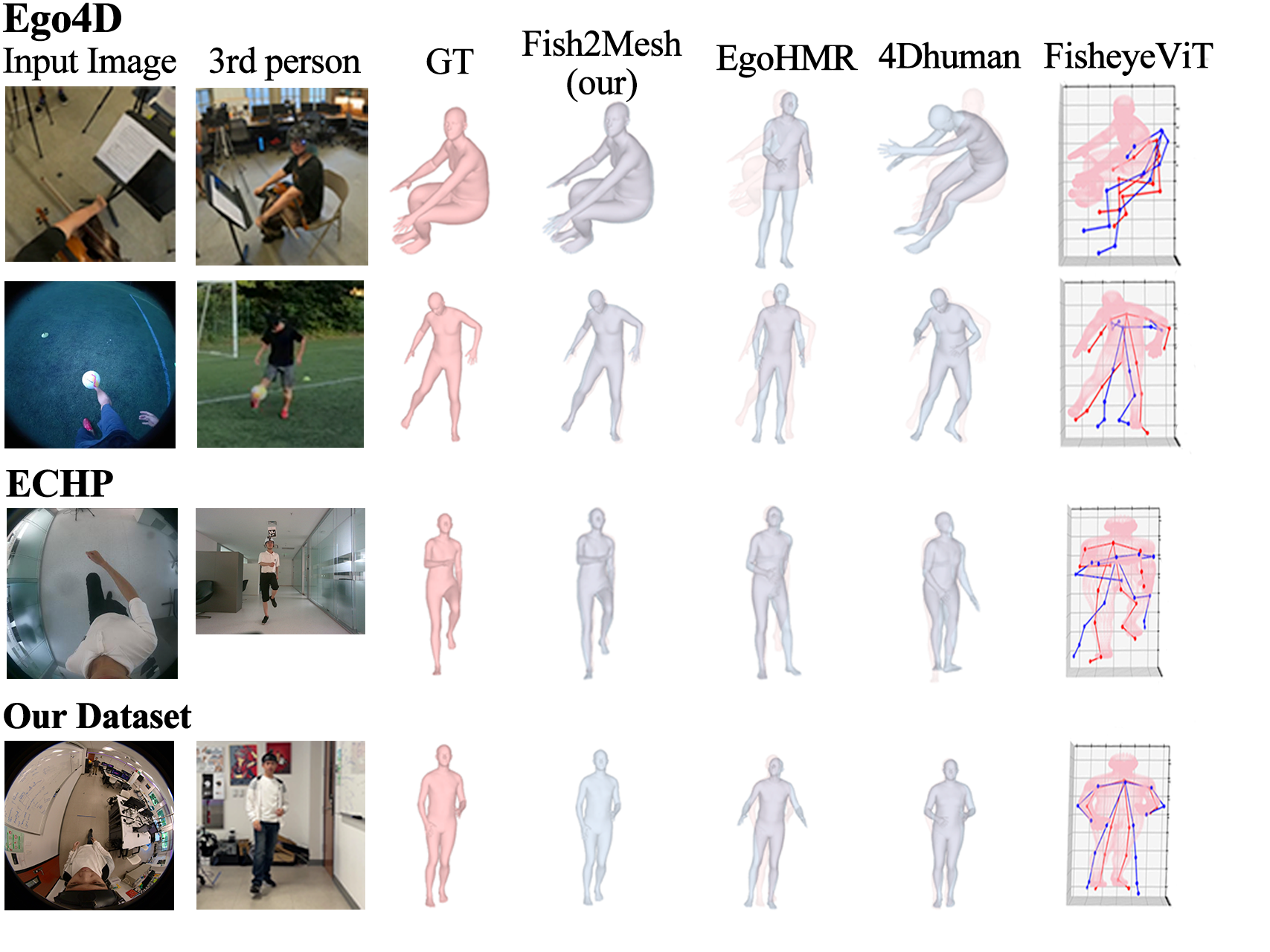} % Reduce the figure size so that it is slightly narrower than the column.
\caption{Visual results of four examples from the four datasets, showing ground truth (red) and related models (blue). FisheyeViT is a pose estimation model, so we visualize the skeleton to compare the resulting joints. The third-person view is not used as model input and is provided purely as an environmental reference.
}
\label{visual}
\end{figure*}

% Additionally, the visualized results demonstrate why it is unfair to compare models using MPJPE and MPVPE across different datasets. 
Our model produces more realistic and accurate human meshes, closely matching the ground truth. Even in challenging cases, such as occlusions caused by body parts or limbs moving out of frame, our model continues to deliver superior performance compared to SOTA models.

As shown in Fig.~\ref{visual}, we selected one example from four datasets to compare the performance of all models. The blue color represents model predictions, while the red color represents the ground truth. Notably, our Fish2Mesh model demonstrates superior estimation of the lower body parts compared to other SOTA models, even when the individual is sitting in a chair or squatting on the ground.

\subsection{Ablation Studies}
% \usepackage{booktabs}
% \usepackage{graphicx}
\begin{table}[]
\caption{Results of ablation study across 3 datasets. Both PA-MPJPE and PA-MPVPE are measured in millimeters (mm).}
\resizebox{\columnwidth}{!}{%
\begin{tabular}{@{}cccc@{}}
\toprule
Model               & PA-MPJPE & PA-MPVPE & Dataset \\ \midrule
Without EPE              & 92.448   & 106.005  & ECHP    \\
Without our dataset & 71.446   & 90.049   & ECHP    \\
Fish2Mesh (our)      & 57.671   & 75.322   & ECHP    \\ \midrule
Without EPE                & 90.111   & 109.995  & Ego4D   \\
Without our dataset & 65.559   & 87.529   & Ego4D   \\
Fish2Mesh (our)      & 41.931   & 54.756   & Ego4D   \\ \midrule
Without EPE               & 87.731   & 99.981   & Our     \\
Without our dataset & 60.446   & 80.782   & Our     \\
Fish2Mesh (our)      & 37.242   & 51.58    & Our     \\ \bottomrule
\end{tabular}%
}
\label{table2}
\end{table}

% To verify the contribution of our ego position embedding and proposed dataset, we conducted two ablation studies.

% In the first study, we removed the ego position embedding from the model. This led to a significant drop in performance, demonstrating that the ego position embedding is crucial for helping the transformer model interpret fisheye images. The embedding enables the model to better understand pixel information, which is vital for accurate pose estimation in such distorted images.

% In the second study, we trained our model using only the ECHP and Ego4D datasets, excluding our proposed dataset. While the performance decline was not as pronounced as in the first study, it still showed a measurable decrease. This confirms that our realistic dataset plays a role in enhancing the model's overall performance.

% For above ablation
To verify the contributions of our EPE and \textbf{proposed dataset}, we conducted two ablation studies, as summarized in Table~\ref{table2}. Each study isolates one of these components to assess its impact on model performance, quantified through PA-MPJPE (Procrustes-Aligned Mean Per Joint Position Error) and PA-MPVPE (Procrustes-Aligned Mean Per Vertex Position Error).

\subsubsection{Removing Egocentric Position Embedding}

As EPE is essential for encoding spherical positional information, our first ablation study removed EPE from the Swin Transformer input of our model and replaced the position embedding to control for its impact on fisheye image distortions. With just a standard position embedding, the model was forced to rely solely on raw image features without any fisheye-specific encoding and lacked structured positional guidance, leading to reduced spatial understanding of the input data.
% (See Table \ref{table2}).
% To assess the contribution of our EPE, we conducted an ablation study where EPE was removed and replaced from the Swin Transformer input, forcing the model to rely solely on raw image features without any fisheye-specific spatial encoding.  

As shown in Table~\ref{table2}, this replacement resulted in a significant drop in performance, with PA-MPJPE increasing from \textbf{57.67 mm} to \textbf{92.45 mm} on the ECHP dataset. This substantial increase in error highlights that EPE enables the model to better interpret pixel information in fisheye images, ensuring accurate depth and positional estimation in such distorted contexts.

\subsubsection{Excluding Proposed Dataset}

In the second study, we trained Fish2Mesh without our proposed dataset, relying solely on the existing Ego4D and ECHP datasets. The proposed dataset contains realistic human activities, varied camera movements, and natural head motions, adding diversity and complexity that enhances the model's generalization to real-world scenarios.

When the model was trained without our dataset, there was a measurable but smaller drop in performance than seen in the first study. Specifically, PA-MPJPE increased from \textbf{57.67 mm} to \textbf{71.44 mm} on the ECHP dataset (Table~\ref{table2}). This decline can be attributed to the fact that the diversity of our augmented ECHP dataset enhanced the model's overall performance.

\section{Conclusion}
In this paper, we introduced Fish2Mesh, a fisheye-aware transformer model tailored for egocentric human mesh reconstruction. Our model addresses three primary challenges in egocentric vision: (1) limited diversity in available datasets for egocentric 3D reconstruction, (2) distortions from fisheye lenses, and (3) self-occlusions due to the egocentric perspective. By leveraging multi-task heads for SMPL parametric regression and camera transformations, Fish2Mesh effectively mitigates these challenges, producing robust, high-fidelity human mesh reconstructions from distorted egocentric input.

Key contributions include a parameterized Egocentric Position Embedding (EPE) that reduces fisheye distortions, as well as a comprehensive training dataset that integrates Ego4D, ECHP, and our newly proposed dataset. This expanded dataset significantly enhances model robustness by capturing diverse scenarios and natural head movements that are common in real-world applications.

The proposed framework demonstrated superior performance over other state-of-the-art methods when evaluated on multiple datasets, including Ego4D, ECHP, and our proposed dataset. In our evaluations, Fish2Mesh achieved the lowest Procrustes-Aligned Mean Per Joint Position Error (PA-MPJPE) and Procrustes-Aligned Mean Per Vertex Position Error (PA-MPVPE) across all tested datasets. These results indicate that Fish2Mesh is not only accurate but also generalizes well across diverse egocentric scenarios, offering a reliable solution for applications requiring high-quality 3D mesh reconstruction from egocentric views.

% \subsection{Future Work}

% The Fish2Mesh Transformer opens up new avenues for research in egocentric human mesh recovery. Future work could focus on further refining the model to handle more complex scenarios, such as dynamic environments and multiple interacting individuals. Additionally, integrating this model with real-time applications in AR/VR and social robotics could significantly enhance user experience and interaction.
% By addressing the inherent challenges of egocentric vision and providing a robust solution for human mesh recovery, our work contributes to the advancement of computer vision and its applications in various fields. We will make our code and dataset publicly available to facilitate further research and development in this area.

{
    \small
    \bibliographystyle{ieeenat_fullname}
    \bibliography{main}
}

\end{document}